# Predicting Preference Flips in Commerce Search

**Samuel Ieong**  Samuel.Ieong@microsoft.com
**Nina Mishra**  ninam@microsoft.com
Microsoft Research, Mountain View, CA 94043

**Or Sheffet**  osheffet@cs.cmu.edu
Carnegie Mellon University, Pittsburgh, PA 15213

## Abstract

Traditional approaches to ranking in web search follow the paradigm of rank-by-score: a learned function gives each query-URL combination an absolute score and URLs are ranked according to this score. This paradigm ensures that if the score of one URL is better than another then one will always be ranked higher than the other. Scoring contradicts prior work in behavioral economics that preference between items depends not only on the items but also on the presented alternatives. Thus, for the same query, preference between items $A$ and $B$ may depend on the presence or absence of item $C$. We propose a new model of ranking, the Random Shopper Model, that allows and explains such behavior. In this model, each feature is viewed as a *Markov chain* over the items to be ranked, and the goal is to find a weighting of the features that best reflects their importance. We show that our model can be learned under the empirical risk minimization framework, and give an efficient learning algorithm. Experiments on commerce search logs demonstrate that our algorithm outperforms scoring-based approaches including regression and listwise ranking.

## 1. Introduction

In web search, an item's relevance to a query is usually absolute (Freund et al., 2003; Joachims, 2002; Burges et al., 2005; 2006; Cao et al., 2007; Dou et al., 2008; Crammer & Singer, 2001). Indeed, ranking algorithms assume the existence of a training set of ⟨query, item⟩ pairs that have been labeled in such an absolute sense, e.g., Perfect, Excellent, Good, Fair or Bad. Further, in interpreting user behavior in click logs, the dominant view is that a user either prefers one item to another or vice versa, but not both. Even in the ranking process, a learned function $f$ takes a query and an item and produces a score. This score induces an absolute ordering between any two items.

In the context of consumer behavior, preference between two items is often dependent on the other items that are shown. In a seminal paper, Amos Tversky showed that user preference between alternatives is relative and comparative: when presented with items $A$ and $B$ alone, users may prefer $A$ to $B$, but when presented with a third alternative $C$, users may flip their preference to $B$ over $A$ (Tversky, 1972). We have found similar examples in the search logs from a commerce search engine (refer to Figure 1). (One example of a commerce search engine is Amazon.) For the search query "paper shredders", when shown $A$, a $20 seven-sheet capacity shredder, vs. $B$, a $50 eleven-sheet capacity shredder, users typically prefer $A$ to $B$. One rationale for this preference is that most users prefer saving $30 at the expense of smaller sheet capacity. However, when $C$, a $95 12-sheet capacity shredder, is shown, users flip their preference to $B$ over $A$. The presence of $C$ causes a change in preference possibly because a shredder that can shred eleven sheets for $50 looks like a bargain compared to one that can shred twelve sheets for $95. This is not a one-off example. We find that about 25% of commerce queries have a product $C$ where users click $A$ more than $B$ in the absence of $C$, and $B$ more than $A$ in its presence (details of experiment omitted).

Such behavior violates a well-known axiom in social choice called *independence of irrelevant alternatives*: the preference between two choices should be independent of context (Arrow, 1950). In this paper, we show that violation of IIA is quite sensible and even explainable. As everyday folklore examples, people often order the second most expensive dish of a menu or the second cheapest wine on the list.

We propose a new model of ranking, called the *Random Shopper Model* (RSM) that allows and explains context-dependent user preferences. RSM's main novelty is in viewing features as Markov chains instead of numeric scores. Products are modeled as vertices on a directed graph, and the weight of an edge denotes the probability that a user transitions from one product to another. Intuitively, the weight from $u$ to $v$ reflects how much better $v$ is than $u$ according to this feature. For example, if the feature is "lower price", then with high probability users will transition to cheaper products, with lower probability users will keep to products of similar price, and with even lower probability users will move to more expensive products.

RSM has a weighting of these Markov chains that captures how important each Markov chain is in the mind of an average user. Our hypothetical shopper starts at a product and repeatedly transitions between items, where in every step she randomly picks one Markov chain according to how important it is and moves to a new product using this chain. In this paper, we give algorithms for learning these weights.

The advantage of RSM is that it ranks in context. The preference between two products is not absolute and may change depending on the other products. Markov processes have the property that the process induced on a subset of items can be very different from the original process over all items. For the paper shredder example, Figure 1 demonstrates how RSM recreates the flip between $A$ and $B$ in the presence of $C$. Ranking functions that assume feature values are absolute scores are inherently incapable of reconstructing such flips. While RSM does not fully capture the shopping process of a user, we believe it constitutes a step towards better user modeling beyond scoring products.

**Contributions** To our knowledge, RSM is the first learning model where the input features are Markov chains. We believe it is a good model when users make a large number of pairwise comparisons, such as online shopping. On the theoretical front, we show that RSM fits into the Empirical Risk Minimization framework— given sufficiently many iid samples, one can learn a hypothesis with bounded error with high probability. We establish a formal bound on the sample complexity in Section 3. Next, we present a general learning algorithm for RSM. Our algorithm is iterative and draws upon the work of Haviv and Van Der Heyden (1984) that bounds the changes in stationary distribution due to changes in the transition matrix. The algorithm is presented in Section 4.

We conduct an empirical evaluation of RSM using data obtained from a commerce search engine. We create a challenging test set consisting only of pairs of preference flips. The test set is such that context-oblivious algorithms cannot achieve over 50% accuracy. We give context-dependent features to existing learning algorithms and show that RSM outperforms these approaches. Our experiment, detailed in Section 5, suggests that RSM is better able to predict preference flips in commerce search.

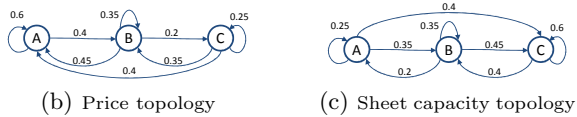

| Product | Price | Sheet Capacity |
|---------|-------|----------------|
| $A$ | $20 | 7 |
| $B$ | $50 | 11 |
| $C$ | $95 | 12 |

(a) Product Specifications

(b) Price topology

(c) Sheet capacity topology

*Figure 1.* The three vertices denote paper shredders with specifications given in (a). We show a topology for "price" in (b) and for "sheet capacity" in (c). Note that roles of $A$ and $C$ are reversed for the two topologies, since $A$ is the cheapest and $C$ has the highest capacity. The weight of the feature "sheet capacity" is 0.4 and of "price" is 0.6. This allows RSM to predict the flip in preference of $A$ over $B$.

## 2. The Random Shopper Model

The Random Shopper Model (RSM) attempts to model the decision process of a user who chooses among a set of items by their features. Under RSM, each feature is viewed as a weighted directed graph called a *topology* (or a *transition matrix* in the algebraic context). A vertex denotes an item and an edge a preference relation. The weight of a directed edge corresponds to the probability that a user transitions from one product to the other. For example, a feature could be "lower price" (Figure 1(b)). The transition probability from $u$ to $v$ increases as $v$ gets cheaper compared to $u$. In addition, there exists a set of weights over these features. A hypothetical user starts with an item and repeatedly performs the following: she picks a feature at random proportional to its weight, and transitions from the current item to another according to the probabilities given by the feature.

Formally, we denote the $i$th topology as $T(i)$, and the weight of each feature by $w^*(i)$. The weights $w^*(i)$ are non negative and sum to 1. (Throughout the paper, all weights and vectors we mention satisfy these two conditions, unless stated otherwise.) We hypothesize that a user follows a random walk according to the combined topology $P(w^*) = \sum_{i=1}^{k} w^*(i)T(i)$, where the weighted sum is computed over the topologies interpreted as transition matrices. The stationary distribution of this walk determines the final ranked order. This model can be interpreted as viewing users as shoppers who go back and forth among items, constantly seeking one that is better than the item they currently consider. Ranking is therefore done according to where more shoppers are likely to be in the limit.

In our model each query-context pair has $k$ topologies uniquely associated with it, whereas the weights $w^*(i)$ remain fixed throughout all samples. This corresponds to users applying the same considerations for price, size, reviews etc. for different sets of TVs for the same query. This is similar to standard machine learning scenarios where each sample has its own features (Markov chains in our case) while the target hypothesis stays fixed throughout.

To ensure that the combined topology converges to a single stationary distribution, we make the common assumption of an "all random" topology – with probability $\lambda$ the shopper transitions into a random item. We assume $\lambda$ is a constant, fixed throughout the paper. This assumption plays an important role in Section 3.

Given a collection of products to rank for a query, ranking proceeds as follows. We (1) restrict the topologies to the products in the collection, (2) renormalize the weights so that outgoing probabilities from each vertex form a probability distribution, (3) weight the restricted topologies according to their importance, (4) compute the stationary distribution of the resulting random walk, and (5) order the products according to this probability distribution.[1]

We now demonstrate that RSM ranks in context. Recall the paper shredder example. In Figure 1, we exhibit two topologies, one for price, and one for sheet capacity. For the same set of weights (0.6 for price and 0.4 for sheet capacity) $A$ is preferable to $B$ in the absence of $C$, but $B$ is preferable to $A$ in the presence of $C$.

**The Learning Problem** The focus of our work is on learning the weights of the features in the proper learning setting. We assume that the features, i.e., the topologies, are given. Designing these topologies requires domain knowledge and may be difficult for certain domains – just as creating features is challenging for machine learning. We leave it as an interesting direction for future work. Each example is composed of a query $q$, a context $C$ (set of products shown to the user), $k$ topologies for this context: $T^{[q,C]}(1), \ldots, T^{[q,C]}(k)$, and a particular product $u \in C$. In the training set, each example is labeled by $p_u^*[q, C]$ the stationary distribution of $u$ under the topology $\sum_i w^*(i) T^{[q,C]}(i)$. Our goal is to learn weights $w$ for these topologies that best approximates $w^*$, namely, s.t. the difference in two labels produced by $w$ and by $w^*$ is smaller than some given threshold $\epsilon$. We believe that the problem of learning under a feature space of topologies is important and of independent interest.

---

[1]This distribution is different from a weighted combination of the stationary distribution of each of the topologies; the latter problem can be learned using existing algorithms.

Formally, we denote $D$ as some distribution over $\{q, C, u\}$, and assume the existence of some oracle that given $q$ and $C$, provides the learner with the $k$ topologies. We also denote by $S$ our training data of $m$ iid samples from $D$.

**Problem 2.1.** *The Random Shopper Problem is $(\epsilon, \delta)$-learnable if there exists an algorithm that for any $D$, gets $m$ iid examples from $D$, and outputs weights $w$ s.t. for any $q, C$, the weights $w$ induce the stationary distribution $p[q, C]$ of the topology $\sum_i w(i) T^{[q,C]}(i)$, and we have that w.p. $\geq 1 - \delta$*

$$err_D(w) \equiv \mathbf{E}_{(q,C,u)\sim D}\left[|p_u[q,C] - p_u^*[q,C]|\right] \leq \epsilon \quad (1)$$

To simplify notation, we will henceforth drop $C$, treating each query-context combination as its own query. We may also drop $q$ when the context is clear.

Observe that Problem 2.1 is more general than what is traditionally required of ranking algorithms. Typically, one requires that if the target ranking noticeably prefers $u$ to $v$, then the hypothesis outputted should also rank $u$ above $v$. In our setting such a requirement translates to correctly ranking $u$ above $v$ if $p_u^*[q] > p_v^*[q] + \Gamma$ for some given $\Gamma$. A solution to Problem 2.1 for $\epsilon = \Gamma/2$ implies a solution to the traditional ranking problem. Thus, we focus on Problem 2.1.

## 3. Sample Complexity

We now consider learning under RSM. First, we show that the learning problem fits in the Empirical Risk Minimization framework—with sufficiently many iid examples, a hypothesis with small error on the sample will be a hypothesis with small error on the true distribution with high probability. Formally, we show:

**Theorem 3.1.** *Let $D$ be any distribution over problem instances $\{q, u\}$. Fix any desired $\epsilon, \delta > 0$. Let $S$ be a sample of size $m = O(\frac{k}{\epsilon^2} \log(\frac{k}{\lambda \epsilon \delta}))$ drawn iid from $D$. Define the true error $err_D(w)$ of a hypothesis $w$ using Eq. (1) and sample error $err_S(w)$ as*

$$err_S(w) = \mathbf{E}_{(q,u) \in_R S}\left[\,|p_u[q] - p_u^*[q]|\,\right] \,.$$

*then with probability $\geq 1 - \delta$, for every $w$, we have*

$$|err_D(w) - err_S(w)| < \epsilon \,.$$

To prove Theorem 3.1, we consider a discretization of the hypothesis space into a $(k-1)$-dimensional simplex, where two adjacent points differ by at most $\epsilon$ in any coordinate. We show next that on this $\epsilon$-grid, the point $\hat{w}$ closest to $w^*$ yields a stationary distribution that differs from the "true" one by at most $\epsilon'$.

**Lemma 3.2** (Main Lemma). *Fix $\epsilon > 0$. For any $w^*$ and $\hat{w}$, it holds that $|\hat{p}_u[q] - p_u^*[q]| \leq k\epsilon/\lambda$ .*

Using this lemma, we sketch the Proof to Theorem 3.1.

*Proof sketch for Theorem 3.1.* Set $\epsilon' = \epsilon\lambda/3k$. Using the union bound and the Hoeffding bound on a sample of $m = O(\frac{k}{\epsilon^2} \log(\frac{k}{\lambda \epsilon \delta}))$ iid examples taken from $D$, we can show that all of the $(k-1)^{1/\epsilon'}$ hypotheses on the $\epsilon'$-grid have roughly the same true error and sample error, i.e., w.p. $\geq 1 - \delta$, all $w$ on the $\epsilon'$-grid have

$$|\text{err}_D(w) - \text{err}_S(w)| < \epsilon/3 \ .$$

Fix any $w$ in the simplex, and denote $\hat{w}$ as its closest grid point. Lemma 3.2 gives that both $|\text{err}_S(w) - \text{err}_S(\hat{w})| < \epsilon/3$ and $|\text{err}_D(w) - \text{err}_D(\hat{w})| < \epsilon/3$. □

As a corollary, if $k$ is a small constant, there exists a polynomial time algorithm for the RSM learning problem by brute force enumeration of the $(\epsilon\lambda/3k)$-grid.

We now prove Lemma 3.2. In what follows, we refer to vectors with non-negative entries that sum to one simply as *distributions*. We start by recalling the definition of the limiting and the fundamental matrix from Markov chain theory.

**Definition 3.3** (Limiting Matrix). *For Markov chain $P$ with stationary distribution $p$, the* limiting matrix *is defined as*

$$P^\infty = \lim_{i \to \infty} P^i = \mathbf{1} \, p^\intercal \ .$$

The matrix $P^\infty$ represents the result of an "infinite" traversal over the transition matrix $P$. It takes in one step any distribution to the stationary distribution.

**Definition 3.4** (Fundamental matrix). *For Markov chain $P$, the* fundamental matrix *is defined as*

$$Z = [I - (P - P^\infty)]^{-1} \ .$$

Recall that for a geometric series with $|x| < 1$, $\sum_{i \geq 0}^\infty x^i$ converges to $(1-x)^{-1}$. Likewise, for $\|P - P^\infty\|_\infty < 1$,

$$Z = I + (P - P^\infty) + (P - P^\infty)^2 + \ldots \ .$$

Hence, one can bound the norm of $Z$ by

$$\|Z\|_\infty \leq \sum_{i \geq 0} \|P - P^\infty\|_\infty^i = 1/(1 - \|P - P^\infty\|_\infty) \ .$$

Further details on the properties of these matrices can be found in Ch. 4 of (Kemeny & Snell, 1969).

Now consider the stationary distributions $p$ and $p^*$ of the Markov Chains $P(w)$ and $P(w^*)$. Let $\Delta$ be the difference $P(w) - P(w^*)$. Let $Z(w^*)$ be the fundamental matrix of $P(w^*)$. It has been shown in (Schweitzer, 1968) and (Haviv & Van Der Heyden, 1984) that

$$(p - p^*)^\intercal = p^\intercal \Delta Z(w^*) \ . \quad (2)$$

We generalize their results as follows.

**Claim 3.5.** *For any distribution $v$ and a scalar $\lambda > 0$, let $M$ be the outer-product $\lambda \mathbf{1} v^\intercal$. Then, if $\sum_{i \geq 0} (P(w^*) - M)^i$ converges,*

$$(p - p^*)^\intercal = p^\intercal \Delta \left( \sum_{i \geq 0} (P(w^*) - M)^i \right) \ .$$

*Proof.* By construction, for any distribution $x$, $x^\intercal M = \lambda v^\intercal$. Hence, $(p - p^*)^\intercal M = \mathbf{0}^\intercal$. Note that both $p$ and $p^*$ are distributions but their difference is not. Now consider the LHS of the above equation.

$$\begin{aligned}
(p - p^*)^\intercal &= p^\intercal P(w) - (p^*)^\intercal P(w^*) \\
&= p^\intercal (\Delta + P(w^*)) - (p^*)^\intercal P(w^*) \\
&= (p - p^*)^\intercal P(w^*) + p^\intercal \Delta \\
\Rightarrow p^\intercal \Delta &= (p - p^*)^\intercal (I - P(w^*)) + \mathbf{0}^\intercal \\
&= (p - p^*)^\intercal [I - (P(w^*) - M)]
\end{aligned}$$

If $\sum_{i \geq 0} (P(w^*) - M)^i$ converges, it is equal to $[I - (P(w^*) - M)]^{-1}$. Multiplying both sides of the equation with this term concludes the proof. □

We now construct a suitable $M$ to use with Claim 3.5. Let $M = \frac{\lambda}{n} J$, where $J$ is the $n \times n$ all-1-matrix. This corresponds to a *random restart* with probability $\lambda$. Consider $Q^* = P(w^*) - M$. As each row of $Q^*$ sums up to $(1-\lambda)$, we have $\|Q^*\|_\infty < 1$, and so the sum $\sum_{i \geq 0} (Q^*)^i$ converges. This yields

$$(p - p^*)^\intercal = p^\intercal \Delta \left( \sum_{i \geq 0} (Q^*)^i \right) = p^\intercal \Delta \left( I - Q^* \right)^{-1} \quad (3)$$

**Corollary 3.6.** $\|p - p^*\|_\infty \leq \frac{\|\Delta\|_\infty}{\lambda}$

*Proof.* Observe that in Eq. (3), multiplication is on the left, whereas operator norms of matrices are defined for the right. So we bound the norm of the transpose, using the fact that for every matrix $\|A\|_\infty = \|A^\intercal\|_1$.

$$\begin{aligned}
\|p - p^*\|_\infty &\leq \|p - p^*\|_1 \leq \|(\sum_{i \geq 0} (Q^*)^i)^\intercal\|_1 \, \|\Delta^\intercal\|_1 \, \|p\|_1 \\
&\leq \sum_i (\|(Q^*)\|_\infty)^i \|\Delta\|_\infty \cdot 1 = \frac{\|\Delta\|_\infty}{1 - (1-\lambda)} \quad \square
\end{aligned}$$

Corollary 3.6 proves Lemma 3.2 immediately, since for $\hat{w}$, the closest grid point to $w^*$, we have that $\|\Delta\|_\infty \leq \sum_i |\hat{w}(i) - w^*(i)| \|T(i)\|_\infty \leq k\epsilon \cdot 1$.

## 4. The RSM Algorithm

Previously, we have shown that as a corollary to Theorem 3.1, one can solve the RSM learning problem by brute force enumeration for small $k$ in theory. In practice, as borne out by our experience in Section 5, the learning problem can be solved much faster using the iterative algorithm described in Figure 2.

The overall strategy of the algorithm is as follows: given a target stationary distribution $p^*$, at each iteration of the algorithm, we make small changes to the current weights $w^s$ such that the difference between the stationary distributions $p^*$ and $p^s$ becomes smaller. The key to understanding our algorithm is the derivation of the objective function Eq. (4).

Let $w^*$ be the weights such that $p^*$ corresponds to the stationary distribution of the transition matrix $P^* = P(w^*)$. At iteration $s$, with weights $w^s$, the algorithm computes the transition matrix $P(w^s)$, the stationary

1. Set $w^0$ arbitrary. Set $s = 0$.
2. For each query $q$: compute $G^s[q] = \sum_i w^s(i) T^{[q]}(i)$, its stationary distribution $p^s[q]$, and its fundamental matrix $Z^{[q]}(w^s) = \left[I - (G^s[q] - \mathbf{1}(p^s[q])^\mathsf{T})\right]^{-1}$.
3. Solve the following convex optimization problem:
$$\text{minimize}_x \quad \sum_q \left[(p^*_{u[q]}[q] - p^s_{u[q]}[q]) - \sum_i x(i)\left((p^s[q])^\mathsf{T} T^{[q]}(i)\ Z^{[q]}(w^s) e_{u[q]}\right)\right]^2 \quad (4)$$
$$\text{s.t.} \quad \forall i, \ -\eta \leq x(i) \leq \eta \quad (\text{formally}, \ -\min\{\eta, w(i)\} \leq x(i) \leq \min\{\eta, 1 - \lambda - w(i)\})$$
$$\sum_i x(i) = 0$$
4. If $\|x\| \leq \epsilon$ then halt. Otherwise set $w^{s+1} = w^s + x$, $s = s + 1$ and repeat from 2.

Figure 2. The RSM Algorithm

distribution $p^s$, the limiting matrix $P^\infty(w^s)$, and the fundamental matrix $Z(w^s)$. From Eq. (2),

$$(p^* - p^s)^\mathsf{T} = (p^*)^\mathsf{T} [P^* - P(w^s)] Z(w^s)$$
$$= (p^s + (p^* - p^s))^\mathsf{T} [P^* - P(w^s)] Z(w^s)$$

Recall that $T(i)$ denote the $i$-th topology. Let $x^s(i) = w^*(i) - w^s(i)$. In (Schweitzer, 1968), it was shown that $(I - [P^* - P(w^s)] Z(w^s))$ is invertible. Hence, rearranging the preceeding, we can show that $(p^* - p^s)^\mathsf{T}$

$$= (p^s)^\mathsf{T} [P^* - P(w^s)] Z(w^s) \cdot (I - [P^* - P(w^s)] Z(w^s))^{-1}$$
$$= (p^s)^\mathsf{T} \Big[\sum_i x^s(i) T(i) Z(w^s)\Big] \cdot \Big[\sum_{n \geq 0} \Big(\sum_i x^s(i) T(i) Z(w^s)\Big)^n\Big]$$
$$= (p^s)^\mathsf{T} \Big[\sum_{n \geq 1} \Big(\sum_i x^s(i) T(i) Z(w^s)\Big)^n\Big]$$

The difference in the $u$-th coordinate, $p^*_u - p^s_u$, equals

$$p^*_u - p^s_u = (p^s)^\mathsf{T} \Big[\sum_{n \geq 1} \Big(\sum_i x^s(i) T(i) Z(w^s)\Big)^n\Big] e_u \quad (5)$$

where $e_u$ is the indicating vector of coordinate $u$. If one can solve the above in closed form, then $w^*$ can be found in one step. Unfortunately, there does not exist an explicit formula for the roots of Eq. (5). Thus, we approximate the RHS by the first term of the sum,

$$p^*_u - p^s_u \approx (p^s)^\mathsf{T} \Big[\sum_i x^s(i) T(i) Z(w^s)\Big] e_u$$

This is a good approximation when $x^s(i)$ is small, as the sum will be dominated by the linear term. In fact, if $\|x^s\|_\infty < \frac{\alpha}{kn}$, we can show that

$$\Big|(p^*_u - p^s_u) - \sum_i x^s(i) \Big((p^s)^\mathsf{T} T(i) Z(w^s) e_u\Big)\Big| \leq \frac{\alpha^2}{1 - \alpha} < \alpha^2$$

To sum up, we apply iterative gradient ascent, where in iteration $s$, we compute $x^s$ s.t. $\|x^s\|_\infty \leq \eta$ for some small constant $\eta$ and s.t. $x^s$ minimizes the above difference, and set the new weights $w^{s+1} = w^s + x^s$. As a sanity check, we ran experiments over synthetic data, applying both the RSM algorithm and the brute-force algorithm which uses $\epsilon$-grid. The weights found by both algorithms were very close. Also, we found that performances are better if we minimize the pairwise difference between two items according to $p^*$ and $p^s$. This requires a change in the optimization routine. Full details are omitted due to space limitations.

## 5. Experiments on Commerce Data

**Overview** Our goal is to evaluate how well one can predict flips in users' preferences due to changing contexts. We focus on flip prediction as this is a hard problem that no previous algorithms in the ranking literature can solve. Each row of our dataset consists of a query, a context (the top five products shown), and the click-through rates (CTRs) of the products. The CTRs are aggregated over users as we are interested in flips in preferences of the population and not of the individual. The task for the algorithms is to predict the CTRs given query and context. A row is treated as five problem instances, one for each product.

To measure performance on flip prediction, we carefully construct the dataset to be composed of pairs of rows where each row in the pair has the same query but a different context, and for which there are two products $A$ and $B$ where $A$ is preferred to $B$ in one and $B$ to $A$ in the other. The algorithms are evaluated on their ability to predict preference flips. In other words, we measure their performance on predicting the relative CTRs between these two products. Note that this is a particularly challenging test because it is loaded with "contradictions". Any algorithm that produces an absolute score for a query-product pair will be correct on one instance and wrong on the other, and cannot achieve an accuracy of $> 50\%$.

**Dataset** We obtain queries and clicks from a commerce search engine from 08/2010 to 02/2011. We focus on queries related to TVs and digital cameras as these are major categories of consumer products where users carefully examine product attributes. We group the data by query and context, where context is defined as the top five products shown for the query. We believe this is a reasonable definition as the top five products are typically visible to the users without scrolling. For each query and context, we count the number of clicks by all users on the five products.

Next, we examine all products surfaced for a query across all contexts. For each pair of products $A$ and $B$, we look for the existence of two contexts where in one $A$ is clicked more often than $B$ and in another $B$ more often than $A$. We consider a context only if there are more than five total clicks, and the difference in clicks between $A$ and $B$ is at least two. This is done to reduce our exposure to spurious clicks. If there are multiple such contexts, we select the two where the preferences expressed are the strongest, i.e., the differences in CTRs between $A$ and $B$ are the largest. Such pairs of instances are added to our dataset. For each run of our experiment, the dataset is randomly split into 80% training and 20% test data. We ensure that paired rows are not split between training and test.

**Features** The features used in this study are Brand, Price, Diagonal Size/Megapixel (depending on TV or camera), Number of Reviews, Average Rating, BM25 and Position of the product in the result set. We select these features as they are visible to users on the result page (with the exception of BM25, which measures how closely a product title matches a query, and hence is quite "visible" too). All features are numeric except for Brand, which we manually map to the range of $\{-1, 0, +1\}$. Reputable brands, such as Samsung and Sony, receive a label of $+1$, while unknown brands received a label of $-1$, and others $0$.

For RSM, we convert each feature into a weighted digraph with self-loops as follows. For each feature, the $n = 5$ products in the context are ordered by feature value, and assigned a rank from 1 to $n$ where high-ranked products are more desired (e.g., cheaper, better brands, more highly rated). The edge weight from product $i$ to $j$ is set to $[n + \mathrm{rank}(j) - \mathrm{rank}(i)]$ and then normalized so that the weights of outgoing edges from each product sum to 1. We choose this encoding as it is simple, scale-invariant and most importantly relative—it depends on all products shown and not on the actual numeric values. This method of constructing topologies differs from the one detailed in Section 2. This is because position bias ranks exactly five items. To avoid the imbalance of $n = 5$ products for the position bias topology while having extremely large $n$ for other topologies, we construct all topologies with five products. The question of how best to encode a feature as a Markov chain is an interesting future research direction.

**Baseline** Our baseline consists of two algorithms: Least Squares (LS) and Listwise Ranking (LR) (Cao et al., 2007). The objective of LS is to learn a weighted combination of features that best predicts the CTR. Learning such a hyperplane is a natural choice as our task of learning CTR is related to regression. Another baseline is Listwise Ranking. We select it as the algorithm trains on lists of choices instead of pairs. In the classical setting, LR assumes each query is associated with a set of URLs with their relevance labels. In our case, relevance is approximated by CTR. The LR algorithm depends on certain parameters. We select the parameters to maximize performance on a validation set sampled from the training data.

**Metrics** We measure performance by the fraction of flips in preferences an algorithm correctly predicts in the test set. Recall that for each pair of rows of data, there are two products $A$ and $B$ where preferences are flipped depending on context. We compare the predicted CTR for these two products under the two contexts, and count the number of times the predicted CTR agrees with the preference. Under this metric, random guessing will have a performance of 50%. Also, any context-oblivious approach that assigns the same score to a query-product pair will have a performance of 50%.

Note that for this experiment, we cannot apply traditional IR metrics such as MAP, MRR, or NDCG. These metrics rely on relevance labels assigned to query-product pairs. Assigning context-oblivious relevance labels to query-product pairs is not consistent with the fact that preference is context dependent.

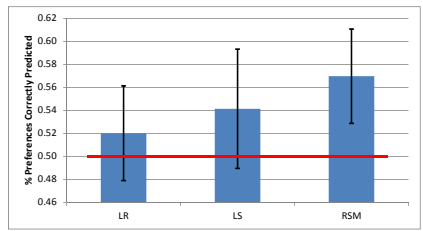

*Figure 3.* Prediction Accuracy of Different Algorithms

**Results** The prediction accuracies of each method averaged over 100 random training-test splits are reported in Figure 3. The solid line at 0.5 in the chart represents the performance of random guessing. The standard deviation of each method is shown in error bars. Despite the overlapping error bars, the differences in performance among all three methods are statistically significant under a paired $t$-test, with $p$-value $< 10^{-5}$. The pairing is done by having all three methods evaluated on the same training-test split. In our experiments, we set $\lambda = 0.15$. Our results are not sensitive to the choice of $\lambda$, from 56.8% with $\lambda = 0.01$ to 57.9% with $\lambda = 0.3$. In our experiments, training time was – RSM: 14s, Listwise:1.3s, LS: 5ms; and ranking time was – RSM: 0.02ms, Listwise: 0.001ms, LS: 0.001ms.

Note that both LR and LS have accuracy $> 0.5$. This is because the Position feature varies depending on where a product is shown. Both LR and LS assigns a negative weight to this feature, predicting that a product ranked higher in the result set is more likely to be clicked, confirming the importance of position bias. While all methods had access to the same features, RSM significantly outperforms both LR and LS. Our novel view of a feature as a Markov Chain, as well as our better modeling of the learning problem, are key to this improved performance.

**Discussion** The experiments have several limitations. First, clicks are a noisy signal for measuring preferences. A click could be due to sheer curiosity or even a mistake. We try to mitigate this problem by requiring a minimum number of clicks. A better signal may be user purchases. Second, all queries are grouped together for learning. It may be that depending on the nature of the query, the weights on the features are different. For example, users who issue [32" LCD TV] may behave differently than those that issue [widescreen TV]. Finally, many important features may be missing, e.g., was the product shown with a photo. Finding important and relevant features remains an ongoing challenge.

In summary, we showed that under a careful setup, beyond statistical doubt, RSM outperforms two strong baselines. Our proof of concept provides real evidence that RSM is a model deserving additional study and experimentation.

## 6. Related Work

**Learning to Rank:** Many techniques have been proposed for the problem of learning to rank, including boosting (Cohen et al., 1999; Freund et al., 2003), gradient descent (Burges et al., 2005; 2006), and large-margin classifiers (Joachims, 2002); see Liu (2009) for a recent survey. Typically, the global ordering is given by some scoring function learned from data. Our work has two important differences. First, we are interested in learning an ordering that depends on the set of items shown to the user, i.e., the context. It is crucial that the method generalize to previously unseen contexts. We do so by learning how the ordering depends on context. Second, our work treats features as topologies over items, whereas most past work deals with standard numeric features. One exception to feature as scores is Cohen et al (1999), where features are viewed as acyclic graphs. Their generalization allows two items to be incomparable. Our work takes the generalization further, allowing each feature to be an arbitrary graph, possibly containing cycles.

Listwise Ranking (Cao et al., 2007) learns from a list of items, and not from pairs of items. It was proposed as a technique to improve computational efficiency. Even though the paper states that each query is associated with a unique list of items, one can view the work in a context-dependent way, i.e., each query is associated with multiple lists of items. However, it still outputs a single value per query-item pair. Xiong et al (2012) observed that the click-through rate (CTR) of an ad is often dependent on the other ads shown alongside, and introduce a context-dependent learning scenario. They discuss algorithms for learning which ad a search engine should surface in response to a query, and propose one that learns from a list of ads rather than one.

**Behavioral Economics:** The foundation of our work is built on the observation that people's preferences are often influenced by context. Some prototypical examples are presented in (Ariely, 2008). The effect of context on preferences has been studied systematically by Tversky. Tversky and Simonson (1993) demonstrated that preferences between two options often depends on other options present. As a consequence, there is no global ranking function that will be consistent with the choices if one ignores the context, which motivates our present work. Tversky (1972) proposed a choice model called *elimination by aspects* (EBA). Under EBA, a decision maker chooses among options by sets of aspects. An example aspect could be {price < $100}. The decision maker chooses an option by picking an aspect and eliminating all choices that do not satisfy the aspect, and repeating until she is left with a single option. In our model, rather than eliminate options, we transition from an inferior option to a superior one along the selected aspect. This process can be viewed as a "softening" of the hard decisions made by EBA.

**PageRank:** Our work can be viewed as a generalization of PageRank (Brin & Page, 1998). Whereas PageRank postulates that users randomly surf from one webpage to another via a hyperlink, our model postulates that users randomly pick a topology according to some distribution, and transition from one item to another based on the selected topology. The distribution over topologies is learned from data such that the stationary distribution closely approximates observed click probabilities. The problem of learning the weights to a random walk has recently been considered by Backstrom and Leskovec (2011). In that paper, the authors study how to assign weights to the edges of a given topology so as to approximate a target stationary distribution. In our problem, topologies and their associated edge weights are given as input, and we are interested in learning the weights of each topology under the aforementioned random walk. In the context of learning, Girolami and Kaban (2004) learn a model of random walks by finding a small set of Markov chains that explain a large collection of transition sequences (they also assume the chains themselves come from some probabilistic model).

**Rank Aggregation:** The problem of using rank aggregation in web search was studied by Dwork et al (2001). Their goal is to aggregate different search results into one ranking that is close to all of the input rankings, where proximity is measured by Kendall tau distance or Spearman footrule (see (Dwork et al., 2001) for definitions). Our goal differs in that we seek to rank results differently depending on the context of the other results shown.

**Using Context to Order:** In the database setting, Agrawal et al (2006) consider the problem of ranking selected tuples in a context-dependent manner. Each context is a conjunction of attributes over a relational table. A collection of preferences is assumed to be given per context, where each preference is of the form attribute value $x$ is preferred to attribute value $y$ in context $Z$. Given a select predicate query, their method finds a ranking of tuples that maximally agrees with the contextual preferences. Our work differs in that we define the context to be the set of items shown to the user. Further, our model supports generalization to contexts, i.e., sets of items, that have not been seen in the input preferences.

## 7. Future Work

We proposed the Random Shopper Model, a new model that can explain contextual preferences in consumer behavior. While this does not directly model how people shop, it moves in a direction that is closer to human behavior than rank by score. It is also a first step towards expanding the view of a feature from a number to a Markov chain. This new view could draw more research interest to non-numeric features such as graphs and Markov chains.

While consumers do flip their preference, characterizing when and why they flip is important. We anecdotally observe that users early in the shopping process are more likely to flip, e.g., flips occur with "steam mop" and not when a precise product is pinned down, e.g., "garmin 265wt". Other reasons for flips include asymmetric dominance and extremal aversion (Tversky & Simonson, 1993). An improved characterization can lead to ML algorithms that prefilter which queries to trigger an algorithm such as RSM vs. triggering the usual ranking algorithm.

Finally, consumer behavior is more complex than predicting flips. The behavioral economics community has studied many other aspects of consumer behavior. Commerce logs open the door to understanding whether and how often such behavior exists. For example, anchoring (Tversky & Kahneman, 1974) suggests that the first product influences subsequent buying decisions, as future products are compared to the first product the user saw. Analogously, in commerce search, the first search result may also have an anchoring effect. Future challenges lie in designing models that better capture consumer behavior.